\documentclass[twoside,11pt]{article}
\usepackage{amsmath}

\usepackage[preprint]{jmlr2e}

\newcommand{\bs}{\boldsymbol}

\usepackage{lipsum}
\newcommand\blfootnote[1]{%
  \begingroup
  \renewcommand\thefootnote{}\footnote{#1}%
  \addtocounter{footnote}{-1}%
  \endgroup
}

\usepackage{lastpage}
\jmlrheading{25}{2024}{1-\pageref{LastPage}}{1/21; Revised 5/22}{9/22}{21-0000}{Yaşar Cahit Yıldırım, Efe Mert Karagözlü, İlter Onat Korkmaz, Çağın Ararat and Cem Tekin}

\ShortHeadings{%
    \texttt{VOP\textnormal{y}}: A Framework for Black-box Vector Optimization
}{Yıldırım, Karagözlü, Korkmaz, Ararat, Tekin}
\firstpageno{1}

\begin{document}

\title{\texttt{VOPy}: A Framework for Black-box Vector Optimization}

\author{%
    \name Yaşar Cahit Yıldırım\textsuperscript{1} \email cahit.yildirim@bilkent.edu.tr\\
    \name Efe Mert Karagözlü\textsuperscript{2*} \email ekaragoz@cs.cmu.edu\\
    \name İlter Onat Korkmaz\textsuperscript{1} \email onat.korkmaz@bilkent.edu.tr\\
    \name Çağın Ararat\textsuperscript{3} \email cararat@bilkent.edu.tr\\
    \name Cem Tekin\textsuperscript{1} \email cemtekin@ee.bilkent.edu.tr\\
    \name \textsuperscript{1} \addr{Department of Electrical and Electronics Engineering, Bilkent University} \\
    \name \textsuperscript{2} \addr{Machine Learning Department, Carnegie Mellon University} \\
    \name \textsuperscript{3} \addr{Department of Industrial Engineering, Bilkent University} \\
}

\editor{My editor}

\maketitle

\begin{abstract}
We introduce \texttt{VOPy}, an open-source Python library designed to address black-box vector optimization, where multiple objectives must be optimized simultaneously with respect to a partial order induced by a convex cone. \texttt{VOPy} extends beyond traditional multi-objective optimization (MOO) tools by enabling flexible, cone-based ordering of solutions; with an application scope that includes environments with observation noise, discrete or continuous design spaces, limited budgets, and batch observations. \texttt{VOPy} provides a modular architecture, facilitating the integration of existing methods and the development of novel algorithms. We detail \texttt{VOPy}'s architecture, usage, and potential to advance research and application in the field of vector optimization.  The source code for \texttt{VOPy} is available at \href{https://github.com/Bilkent-CYBORG/VOPy}{\texttt{https://github.com/Bilkent-CYBORG/VOPy}}.
\end{abstract}


\section{Introduction}
\blfootnote{\textsuperscript{*}Part of this work was done when the author was at Bilkent University.}

Black-box optimization is a subfield of optimization where the optimal value of a function is sought without using derivatives, primarily relying on potentially noisy pointwise evaluations of the function. In many real life problems, gradients are not available, making black-box optimization methods invaluable. These applications include hyperparameter tuning of machine-learning algorithms \citep{wu2019hyperparameter, snoek2012practical}, neural architecture search \citep{kandasamy2018neural}, portfolio allocation \citep{hoffman2011portfolio}, and synthetic gene design \citep{gonzalez2015bayesian}. Black-box optimization is also widely used in experimental design \citep{terayama2021black}, such as thermal conductivity \citep{seko2015prediction} and crystal-structure prediction \citep{oganov2006crystal}. Libraries like \texttt{COMBO} \citep{ueno2016combo}, \texttt{RBFOpt} \citep{costa2018rbfopt}, \texttt{Google Vizier} \citep{golovin2017google}, and \texttt{OpenBox} \citep{jiang2024openbox} have been implemented to perform black-box optimization.

Despite the success of single-objective optimization, many applications require simultaneous optimization of different objectives. Multi-objective optimization (MOO) usually deals with this using the componentwise order for objective vectors \citep{emmerich2018tutorial}. However, one may want to consider a more general ordering to reflect more and less conservative ways in which one design dominates another. In vector optimization (VO), one defines a vector partial order $\preceq_C$ on $\mathbb{R}^D$ induced by a closed convex cone $C$ with $C\cap -C=\{\bs 0\}$ via $\bs \mu \preceq_C \bs \nu$ if and only if $\bs \nu - \bs \mu \in C$. Here, $D\geq 2$ is the number of objectives and $\bs \mu,\bs \nu\in\mathbb{R}^D$ are arbitrary objective vectors. Then, the goal is to find the \textit{Pareto set} of a vector-valued function $\bs f$ with respect to $\preceq_C$ over a given set of designs through sequential pointwise observations. In different applications, there could be observation noise, limited observation budget, partial observations of $\bs f$, an ability to accommodate batch observations, etc. Although there is recent focus on black-box vector-optimization algorithms \citep{ararat2023vector, karagozlu2024learning, korkmaz2024vector}, a unifying and convenient framework covering these cases is missing for algorithm designers and practitioners. Here, we introduce \texttt{VOPy} to facilitate the development and applications of the field.

\noindent \textbf{Related work.}
While there are no tools specifically designed for black-box VO, several libraries have been developed for MOO. An important example is \texttt{PyMOO} \citep{blank2020pymoo}, which offers a comprehensive suite of ready-to-use MOO algorithms. However, it does not include Bayesian methods and falls short in facilitating decision-making processes crucial for end-to-end black-box algorithms. Some tools take advantage of Bayesian optimization (BO), which is widely used in MOO applications. For instance, \texttt{PyePAL} \citep{jablonka2021bias} implements the \texttt{$\epsilon$-PAL} algorithm \citep{zuluaga2016pal}, providing utilities for its application across various problems. However, it lacks flexibility for adapting to new or varied algorithms. \texttt{OpenBox} \citep{jiang2024openbox}, a comprehensive platform for black-box optimization, supports MOO. However, it lacks the modularity needed to accommodate custom algorithm development. Lastly, \texttt{BoTorch} \citep{balandat2020botorch} is a well-rounded library that is highly regarded for its extensiveness and modularity in Bayesian optimization. It has additional support for decoupled algorithms, a feature shared with \texttt{VOPy}. While these MOO libraries offer various capabilities, they all lack support for the generalized partial orderings essential for vector optimization—a gap \texttt{VOPy} is designed to address.

\noindent \textbf{Contributions.}
\textbf{(1)} We design and implement \texttt{VOPy}, the first library for developing black-box vector optimization algorithms.
\textbf{(2)} We provide solutions to the convex optimization problems that arise while performing confidence region comparisons with respect to $\preceq_C$.
\textbf{(3)} We implement some of the existing VO and MOO algorithms from the literature using the modular design of \texttt{VOPy}.
\textbf{(4)} We provide support for decoupled algorithms, which enable partial evaluations when objective costs may differ, along with novel example algorithms.

\section{\texttt{VOPy}} \label{sect:VOPy}

\begin{figure}[ht]
    \centering
    \includegraphics[width=0.7\linewidth]{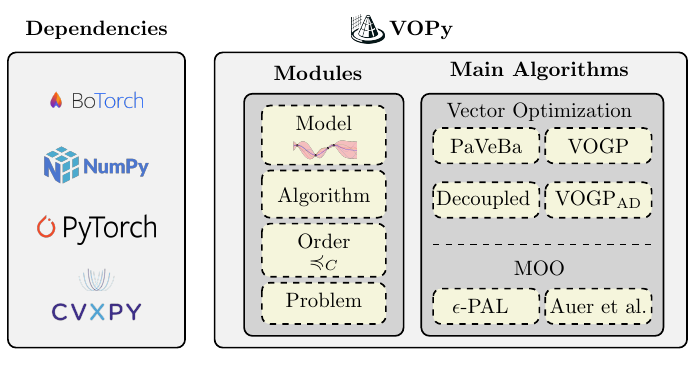}
    \caption{Overview of the dependencies, core modules, and built-in algorithms of \texttt{VOPy}.}
    \label{fig:overview}
\end{figure}

\texttt{VOPy} is an open-source library where it is possible to use the existing vector optimization methods for applications, develop new algorithms using the efficient framework of \texttt{VOPy} for uncertainty and partial-order modeling, and to develop methods almost from scratch and benchmark them against known methods. \texttt{VOPy} adopts a modular design philosophy, inspired by the Bayesian optimization library \texttt{BoTorch}. An overview of \texttt{VOPy} with its dependencies, core modules, and built-in algorithms can be seen in Figure~\ref{fig:overview}.

\subsection{Design and Implementation}

The architecture of \texttt{VOPy} is divided into four main interfaces: \texttt{Order}, \texttt{Model}, \texttt{Algorithm}, and \texttt{Problem}. The \texttt{Order} interface is central to defining the partial order used across the library. This can be the componentwise order in MOO or any other partial order defined by a polyhedral ordering cone. This interface is critical as it allows all components of the library to perform solution comparisons.
The \texttt{Model} interface facilitates the creation of various modeling components, which can use frequentist or Bayesian methods. \texttt{VOPy} comes with some built-in models that should suffice for most quick implementations: the empirical mean and variance model, multi-output Gaussian processes (GP) with independent or coregionalized outputs, and GPs that support decoupled modeling.
The \texttt{Algorithm} interface is designed to foster algorithms. Implementations of the state-of-the-art VO algorithms as well as some important MOO algorithms are presented together with \texttt{VOPy}.
Lastly, with the \texttt{Problem} interface, the user can connect their optimization problem with the other components of \texttt{VOPy}. Necessary tools to generate problems from offline datasets and link up real world problems are available.
Together, these interfaces promote a flexible yet structured approach for developing and implementing end-to-end black-box VO algorithms, ensuring that \texttt{VOPy} can be adapted to a wide range of optimization problems and settings.
In noisy black-box optimization, possible solutions have confidence regions containing the true function value with high probability.
In MOO, hyperrectangles are used which fits well with componentwise order for easy comparison between regions.
However, this approach does not extend to arbitrary partial orders, nor can it exactly represent models involving dependent objectives.
To address this, we provide a \texttt{ConfidenceRegion} abstract class with \texttt{RectangularConfidenceRegion} and \texttt{EllipsoidalConfidenceRegion} derivatives, both supporting solutions to these comparisons using the \texttt{CVXPY} library.

\texttt{VOPy} is designed with performance in mind and provides significant speed-ups when possible. (see \href{https://vopy.readthedocs.io/en/stable/examples/check_dominates_performance.html}{Performance notebook}). \texttt{VOPy} follows the coding conventions of PEP8 \citep{pep8} and enforces it through Flake8, Black and $\mu$sort. \texttt{VOPy} also has been checked for vulnerabilities with the tool Bandit and is secure. These choices align with most of the frequently used libraries for Python community, e.g., the PyTorch ecosystem.

\subsection{Testing and Accessibility}

\texttt{VOPy} is extensively tested with a test suite that covers a broad range of code functionality. By maintaining a high code coverage ($\geq 95\%$), \texttt{VOPy} ensures reliability for both users and developers. Documentations for \texttt{VOPy} are generated using the Sphinx autodoc, and are hosted on ReadTheDocs, accessible through our GitHub repository.

\section{How to use \texttt{VOPy}?} \label{sect:howtouse}
\texttt{VOPy} includes several pre-implemented algorithms, models, orders, and problems from the literature for black-box vector optimization, allowing users to select and use components based on their specific needs. The library provides various VO algorithms, including Naive Elimination \citep{ararat2023vector}, PaVeBa and PaVeBa-GP \citep{karagozlu2024learning}, and VOGP together with its extension to continuous domains using adaptive discretization \citep{korkmaz2024vector}. From MOO literature, \texttt{VOPy} implements Algorithm~1 of \citep{auer2016pareto}, which has no official code released, and $\epsilon$-PAL \citep{zuluaga2016pal}.
It also includes a decoupled version of PaVeBa and a novel entropy-based decoupled algorithm.

\texttt{VOPy} natively supports componentwise orders, 2D cones, polyhedral approximations of 3D ice-cream cones,
and polyhedral cones defined by a coefficient matrix. The library includes various GPyTorch models that handle coupled, decoupled, dependent, and independent objectives. For real-world datasets with vector-valued outputs, \texttt{VOPy} provides the SNW dataset \citep{zuluaga2012computer} and datasets for the Disk Brake \citep{tanabe2020easy} and Vehicle Safety \citep{liao2008multiobjective} problems, generated using Sobol sequences.

\noindent \textbf{Using existing methods for new problems.}
Researchers and practitioners can define their specific problem instances and integrate their experimental data to apply \texttt{VOPy}’s black-box vector-optimization tools. These ready-to-use components can be directly applied to optimization tasks with minimal setup. Following the documentation’s examples, users only need to connect their problem through a dataset or problem structure, set up the appropriate algorithm with an existing model and order, and let the algorithm run.

\noindent\textbf{Benchmarking.}
Algorithm developers can create new algorithms by inheriting from \texttt{VOPy}’s abstract algorithm class, leveraging existing models, orders, and problems to compare their results with existing algorithms. \texttt{VOPy}’s benchmarking capabilities include metrics such as $\epsilon$-F$1$ score \citep{karagozlu2024learning} and hypervolume discrepancy \citep{tu2022joint}.

\noindent\textbf{Custom uses.}
Any module in \texttt{VOPy} can be replaced with custom methods, e.g., users can incorporate a novel uncertainty model, using it together with existing algorithms and orders. Alternatively, if an application requires a non-polyhedral cone, 
the practitioner can define a custom \texttt{Order} derivative and apply existing algorithms and models on the problem.

\section{Results and Conclusions}
\texttt{VOPy} is the first open-source library designed for black-box VO.
Its early stages contributed to the development of algorithms now existing in the literature \citep{karagozlu2024learning, korkmaz2024vector}.
\texttt{VOPy} offers compatibility with various orders, models, algorithms, and acquisition strategies.
With its demonstrated reliability and ease of use, it enables users to apply built-in tools or integrate new methods. In summary, \texttt{VOPy}
offers a robust platform with the potential to lead further research and applications in black-box VO.


\acks{%
This work is supported by T\"{U}B\.{I}TAK grant 121E159.
Y.C.Y. and İ.O.K. are supported by Türk Telekom as part of 5G and Beyond Joint Graduate Support Programme coordinated by Information and Communication Technologies Authority.
C.T. acknowledges support from a T\"{U}BA-GEB\.{I}P 2023 Award. 
}




\vskip 0.2in
\bibliography{references}

\end{document}